\documentclass[dvipsnames]{article}

\usepackage[final]{neurips_2024}

\usepackage{tabularx} % 放在导言区
\usepackage{amsmath}
\usepackage{amssymb}
\usepackage{mathtools}
\usepackage{amsthm}
\usepackage{multirow}
\usepackage{makecell}
\usepackage{caption}
\usepackage{subcaption}
\usepackage{wrapfig}

\usepackage{graphicx}
\usepackage{pifont}
\usepackage{textcomp}

\usepackage[utf8]{inputenc} % allow utf-8 input
\usepackage[T1]{fontenc}    % use 8-bit T1 fonts
\usepackage{url}            % simple URL typesetting
\usepackage{booktabs}       % professional-quality tables
\usepackage{amsfonts}       % blackboard math symbols
\usepackage{nicefrac}       % compact symbols for 1/2, etc.
\usepackage{microtype}      % microtypography

\usepackage{color, soul}
\usepackage[most,breakable]{tcolorbox}
\usepackage{xcolor}         % colors

\usepackage[labelfont=bf]{caption}
\usepackage{enumitem}
\usepackage{sectsty}

\usepackage[colorlinks, linkcolor=RoyalBlue, anchorcolor=BrickRed, citecolor=RoyalBlue, urlcolor=RoyalBlue]{hyperref}
\usepackage{amsthm}

\title{H-Neurons: On the Existence, Impact, and Origin of Hallucination-Associated Neurons in LLMs}

\author{
Cheng Gao,
Huimin Chen, Chaojun Xiao,  Zhiyi Chen, 
Zhiyuan Liu, Maosong Sun\\ 
    Tsinghua University\\
 \texttt{\{gaoc24\}@mails.tsinghua.edu.cn}, \texttt{\{huimchen,xcj,liuzy\}@tsinghua.edu.cn}
}

\begin{document}

\maketitle
\vspace{3em}

\begin{abstract}
Large language models (LLMs) frequently generate hallucinations -- plausible but factually incorrect outputs -- undermining their reliability. While prior work has examined hallucinations from macroscopic perspectives such as training data and objectives, the underlying neuron-level mechanisms remain largely unexplored. In this paper, we conduct a systematic investigation into hallucination-associated neurons (H-Neurons) in LLMs from three perspectives: identification, behavioral impact, and origins. Regarding their identification, we demonstrate that a remarkably sparse subset of neurons (less than $0.1\%$ of total neurons) can reliably predict hallucination occurrences, with strong generalization across diverse scenarios. In terms of behavioral impact, controlled interventions reveal that these neurons are causally linked to over-compliance behaviors. Concerning their origins, we trace these neurons back to the pre-trained base models and find that these neurons remain predictive for hallucination detection, indicating they emerge during pre-training. Our findings bridge macroscopic behavioral patterns with microscopic neural mechanisms, offering insights for developing more reliable LLMs.
\end{abstract}

% \keywords{keyword1, Keyword2, Keyword3, Keyword4}

\section{Introduction}\label{sec1}
% 近年来，大型语言模型在自然语言处理任务中取得了突破性进展，展现出强大的语言理解和生成能力。然而，伴随着这些突破性进展,一个关键的可靠性问题始终困扰着研究者和使用者：幻觉(Hallucination)。幻觉是指模型生成看似流畅合理但实际上错误或无根据的内容，这种现象严重削弱了用户对模型输出的信任。例如，xxxxx（加一些在不同领域的幻觉研究 @gaocheng）

% 为了提升大模型的可靠性，许多研究者投入了大量努力去探索并理解大模型幻觉的产生机制及其影响因素，大体可分为三类：1）从训练数据来看，数据中的不均衡分布和内在偏差导致模型难以可靠记忆罕见事实。（列举几个具体的研究）。2）从训练目标来看，模型预训练与后训练仅鼓励模型产生正确答案，而未鼓励模型在不确定知识上表达不确定性，导致模型期望冒险输出错误的猜测。例如预训练过程中下一词预测目标优化了续文的似然性而非真实性，指令微调或RLHF鼓励模型生成看似正确的回答而非表达不确定。3）从解码算法来看，生成过程中的随机性和自回归生成带来的错误累积使得输出不稳定，易于从细微偏差演变为幻觉。

% 这些研究将大模型当成一个黑盒系统，在宏观层面对大模型幻觉产生的原因进行了分析，却忽视了神经元层面的微观机制理解。神经元级别的微观分析对揭示幻觉产生机理与大模型幻觉治理具有重要潜力。正如，生物学中细胞分裂机制的研究推动了对癌症等疾病的宏观模型构建和治疗策略的开发；脑科学中单个神经元的电活动和突触交互，促进了对学习、记忆等认知行为的宏观理论建立。神经元是大模型计算的基本单元，观察其激活行为与幻觉之间的关联可以加深对模型可靠性的理解。在可解释性层面，神经元级别的分析可以揭示模型何时产生幻觉；在模型行为控制与对齐层面，神经元提供了一种可干预的杠杆，如通过激活或抑制少量神经元来改变行为。

% 在这篇文章中，我们从神经元视角出发，探讨幻觉在大型语言模型中的微观机制。已有研究发现大模型内部的隐藏状态为幻觉检测提供有效的特征输入，Anthropic等研究利用稀疏自编码器（SAE）为幻觉与神经元激活的关联提供了案例分析，这表明内部神经元的激活行为与幻觉的产生具有潜在关系。进一步地，我们系统性地研究了幻觉相关神经元的存在性、行为机制及其起源。具体而言，本文follow了xxxx的设定，聚焦探索了Feedforward Network中的神经元，并且关注的是在知识类问答上的幻觉表现。我们关注以下三个研究问题：1）Do H-Neurons exist? xxxx 2）How do these neurons influence model behavior? xxxx 3）When do these neurons originate? xxxx
In recent years, large language models (LLMs) have achieved groundbreaking advancements in natural language processing tasks, demonstrating impressive potential towards artificial general intelligence~\citep{foundation-model,gpt3,instruct-gpt,gpt4}. However, these advancements come with a persistent reliability challenge that troubles researchers and users alike: hallucinations. Hallucinations occur when models produce outputs that seem plausible but are factually inaccurate or unsupported by evidence~\citep{OnFaithfulnessandFactuality,hallusurvey}. For example, GPT-3.5 has been shown to hallucinate in approximately 40\% of citation-based factuality evaluations, a figure that improves but remains high at 28.6\% for GPT-4~\citep{Hallucination_Rates}. Similarly,  emerging reasoning-centric systems such as DeepSeek-R1, despite demonstrating strong performance on complex tasks, continue to exhibit pronounced hallucination modes \citep{vectara_deepseek_r1_hallucination}. Collectively, these observations indicate that hallucinations persist regardless of model architecture, highlighting a critical bottleneck in the reliability of state-of-the-art LLMs.

 % Even GPT-5, which reports substantial improvements in factual consistency, still exhibits non-trivial hallucination behaviours according to independent analyses \citep{openai_gpt5_systemcard,nature2025_gpt5}.

To improve LLM reliability, researchers have invested considerable effort in uncovering the mechanisms and factors behind hallucinations, which can be broadly grouped into three categories. First, from a training data perspective, distribution imbalances and inherent biases within datasets make it difficult for models to accurately recall long-tail facts~\citep{DBLP:conf/naacl/SunXZLD24,DBLP:conf/acl/LiLSDSLJJL22}. 
Second, training objectives in both pretraining and post-training phases primarily incentivize confident predictions without promoting the expression of uncertainty for unfamiliar information, encouraging models to output incorrect guesses~\citep{DBLP:journals/corr/abs-2509-04664}. Specifically, the next-token prediction goal in pretraining prioritizes fluent continuations over factual accuracy, while instruction tuning or reinforcement learning often favors generating superficially helpful responses, sometimes at the expense of honest refusals to answer. 
Third, decoding algorithms introduce instability through randomness and error accumulation in autoregressive generation, allowing small deviations to snowball into hallucinations~\citep{DBLP:conf/icml/ZhangPMLS24,DBLP:conf/nips/LeePXPFSC22,DBLP:conf/nips/KapoorGRCPBWDGW24}.

Current studies largely treat LLMs as black boxes, examining hallucination causes at a macroscopic level while neglecting microscopic insights into neuron-level mechanisms. Yet, such fine-grained analysis holds immense promise for explaining how hallucinations arise and for developing mitigation strategies. 
Just as biological research on cellular division informs treatments for diseases such as cancer~\citep{collins1997cell,matthews2022cell}, and neuroscience investigations into individual neuronal activity and synaptic interactions shape theories of cognition like learning~\citep{DBLP:journals/natmi/LuczakMK22} and memory~\citep{mongillo2008synaptic,lisman2018memory}, analyzing neurons -- the fundamental computational units of LLMs -- is essential for decoding hallucination. By scrutinizing neurons' activation patterns in relation to hallucinations, we can gain deeper insights into model reliability. In terms of interpretability, neuron-level analysis can enable the prediction of when hallucinations are prone to emerge; for alignment and behavioral control, it provides actionable intervention points, such as activating or suppressing specific subsets of neurons to reliably modify model outputs.
% 其中三篇生医方向论文似乎没有dblp格式的引用
% Much like how biological research on cellular division has informed macroscopic models and treatments for diseases such as cance这个不是偏宏观嘛需要放在这里说么？

In this paper, we adopt a neuron-centric perspective to investigate the microscopic mechanisms of hallucinations in LLMs. Prior research has shown that internal hidden states can serve as effective features for detecting hallucinations~\citep{Internal_States}, and others using sparse autoencoders have provided case studies connecting hallucinations to specific neuron activations~\citep{anthropicSAE,SAEhallu}, hinting at a deeper link between neuronal behavior and hallucination generation. 
Building on this foundation, we identify a set of hallucination-associated neurons and term them as \textbf{H-Neurons}. We then systematically explore the existence, behavioral impacts, and origins of H-Neurons. We address the following three research questions:
% Building on this foundation, we systematically explore the existence, behavioral impacts, and origins of hallucination-associated neurons. We address the following three research questions:
\begin{itemize}[topsep=1pt,leftmargin=*]
    \item \textit{Q1: Do H-Neurons exist?} Can we identify specific neurons whose activations reliably distinguish between hallucinatory and faithful outputs?
    \item \textit{Q2: How do these neurons influence model behavior?} Specifically, what types of tasks exhibit a significant change on performance when these neurons' activations are altered, thereby establishing a link between hallucination and another capability?
    \item \textit{Q3: When do these neurons originate?} Are they introduced during the post-training alignment phase or already present in the pre-trained phase?
\end{itemize}

Specifically, drawing from setups in previous work~\citep{Finding_Safety_Neurons,Finding_Skill_Neurons,Detecting_hallu}, we focus on neurons in the feedforward networks and examine hallucinations in knowledge-based question answering and make the following observations.

\textbf{Existence of H-Neurons}\quad
% 我们首先将幻觉神经元定义为激活值能够用于准确预测模型是否产生幻觉的神经元。为了发现这些H-Neurons，我们
Our investigation reveals that a remarkably sparse subset of neurons -- comprising less than $0.1\%$ of the model's total neurons -- can accurately predict whether the model will produce hallucinated responses. We refer to these predictive neurons as \textit{H-Neurons}.
To identify these neurons, we develop a systematic methodology that contrasts activation patterns between faithful and hallucinated responses, then apply sparse logistic regression to uncover the most predictive neurons. Notably, the neurons identified through simple QA tasks demonstrate strong generalization capability: they maintain robust predictive accuracy across out-of-distribution scenarios, ranging from specialized cross-domain contexts to pure fabrications concerning non-existent entities, achieving reliable hallucination detection.

\textbf{Impact on Model Behavior}\quad
Our analysis uncovers that H-Neurons are linked to \textit{over-compliance} behaviors in LLMs. To establish this causal relationship, we conduct controlled interventions by systematically scaling the activation magnitudes of these neurons. The interventions reveal a distinctive behavioral pattern: amplifying H-Neurons' activations systematically increases a spectrum of over-compliance behaviors -- ranging from overcommitment to incorrect premises and heightened susceptibility to misleading contexts, to increased adherence to harmful instructions and stronger sycophantic tendencies. These findings suggest that H-Neurons do not simply encode factual errors, but rather represent a general tendency to prioritize conversational compliance over factual integrity.
% Second, we establish the causal role of these neurons through targeted interventions. Scaling their influence within the model reveals consistent behavioral shifts: while accuracy on knowledge-intensive QA tasks remains stable, increased activation of hallucination neurons systematically amplifies behaviors such as overcommitment to incorrect premises, susceptibility to misleading contexts, harmful instruction-following, and sycophancy. This suggests that hallucination neurons encode a broader disposition towards \textit{compliant answer generation}, even when fidelity is compromised.

\textbf{Origin of H-Neurons}\quad
Our investigation reveals that H-Neurons originate during the pre-training phase, providing empirical evidence for the insights proposed by OpenAI researchers from the perspective of learning theory~\citep{DBLP:journals/corr/abs-2509-04664}. To trace their developmental timeline, we conduct cross-model transfer experiments: we apply the hallucination neurons identified in instruction-tuned models to their corresponding base models and evaluate their predictive efficacy. The results demonstrate that these neurons retain their predictive ability in base models, successfully detecting hallucinations even prior to fine-tuning.

% Together, these findings provide the first neuron-level evidence that hallucinations are not diffuse model artifacts but are mediated by a sparse, identifiable set of units—units that simultaneously encode broader answer-generation dispositions such as compliance and overcommitment. Our work advances the mechanistic understanding of hallucinations and introduces concrete levers for future mitigation strategies in model alignment.
% Figure~\ref{fig:framework} presents the overall framework.

In summary, this paper provides a systematic neuron-level investigation into the microscopic mechanisms of hallucinations in LLMs. By bridging the gap between macroscopic behavioral patterns and fine-grained neural activations, we hope our work can deepen the understanding of how hallucinations arise at the computational level, and offer actionable insights for developing more reliable LLMs.

\section{Identification of H-Neurons}\label{sec:Identification}

% 每个部分，先写方法，再写results吧，这样会比较顺畅。方法的介绍，尽可能简短

% 在这一个section中，我们首先关注大模型中是否存在于幻觉相关的神经元。
While prior work has demonstrated that internal hidden states can detect hallucinations~\citep{Internal_States,anthropicSAE,SAEhallu}, a systematic investigation into hallucination-associated neurons remains absent. In this section, we address our first research question: \textit{Do H-Neurons exist?} We hypothesize that among the millions of neurons in modern LLMs, a sparse subset exhibits activation patterns that systematically distinguish between hallucinatory and faithful outputs. The sparse subset of neurons could serve as both interpretable indicators for detection and precise intervention points for further research.

\begin{figure}[t]
    \centering
    \includegraphics[width=\linewidth]{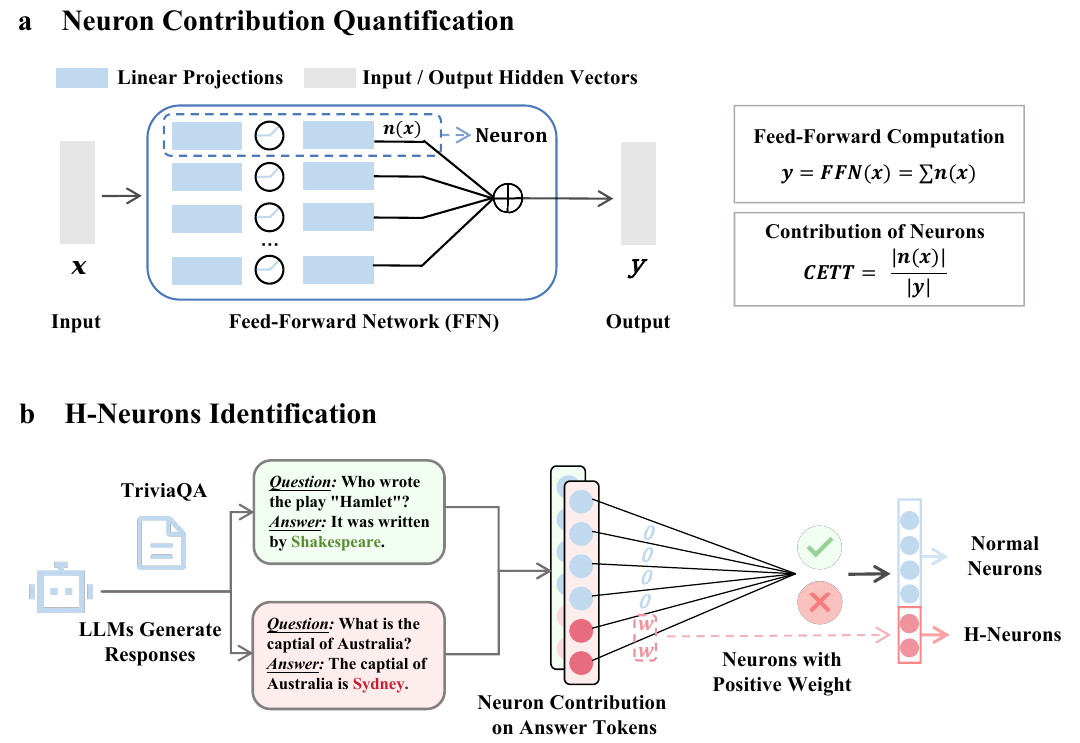}
    % \vspace{-5pt}
    \caption{
        \textbf{Framework for identifying H-Neurons.} 
        (a) Within the Feed-Forward Network, we calculate the contribution of each neuron in one forward pass. This metric normalizes the magnitude of an individual neuron's projected output $|n(x)|$ against the layer's total output vector $|y|$, providing a standardized measure of its contribution to the hidden state.
        (b) The process begins by generating a balanced dataset of faithful (green check) and hallucinatory (red cross) responses using the TriviaQA benchmark. We extract the contribution profiles of neurons specifically on the \textit{answer tokens} to train a linear classifier. Neurons assigned \textbf{positive weights} by this classifier are identified as "H-Neurons", distinguishing them from normal neurons based on their predictive role in generating hallucinations.
    }
    \label{fig:framework_1}
\end{figure}

To identify H-Neurons from the vast parameter space of LLMs, we employ a sparse linear probing approach (Figure~\ref{fig:framework_1}). We first quantify each neuron's contribution to the responses using the CETT metric~\citep{relu2wins}, which is used to measure the neuron's activation level during generation. 
We then frame hallucination detection as a binary classification problem: predicting whether a response is hallucinatory based on neuron activations. Using logistic regression with L1 regularization, we train a sparse classifier that automatically selects the most predictive neurons by driving most weights to zero. 
The neurons with non-zero weights are identified as H-Neurons. Training data is collected from TriviaQA~\citep{Triviaqa} by sampling multiple responses per question and labeling them based on factual correctness. To present the effectiveness of H-Neurons, we establish a baseline by training linear classifiers using randomly selected neurons. To ensure the fairness of the comparison, the number of randomly selected neurons is the same as that of H-Neurons.

% \begin{table}[h]
% \centering
% \caption{Illustrative examples of two distinct hallucination types: (1) questions the model has likely seen during pretraining but answers incorrectly (Seen-but-wrong), and (2) questions about nonexistent entities where the model nonetheless generates confident answers (Unseen-but-answering).}
% \begin{tabular}{p{0.3\textwidth} p{0.2\textwidth} p{0.4\textwidth}}
% \toprule
% \textbf{Question} & \textbf{Model Response} & \textbf{Dataset \& Type} \\
% \midrule
% Who discovered penicillin? & Alexander Graham Bell & TriviaQA (Seen-but-wrong) \\
% Who manufactures the medicine volor pri octacap? & Sun Pharma & NonExistentRefusal (Unseen-but-answering) \\
% \bottomrule
% \end{tabular}
% \label{tab:hallucination-examples}
% \end{table}

\begin{table}[b]
    \centering
    \small
    \vspace{-1.5em}
    \caption{Hallucination detection accuracy ($\%$) of neuron-based classifiers. We evaluate the performance of neuron-based classifiers on six widely used LLMs. Here, "Random" and "Hallucination" refer to classifiers trained with randomly selected neurons and H-Neurons. Ratio refers to the proportion of total neurons that are selected for classifiers. Classifiers with H-Neurons can effectively detect hallucination for in-domain questions~(TriviaQA and NQ), cross-domain questions~(BioASQ), and fabricated questions, demonstrating robustness of H-Neurons. H-Neurons usually account for less than 1\textperthousand{} of all the neurons in LLMs.}
    \begin{tabular}{l|cc|cccc}
    \toprule
    Models  & Neurons & Ratio~(\textperthousand) & TriviaQA & NQ-Open & BioASQ & NonExist   \\
    \midrule
    \multirow{2}{*}{Mistral-7B-v0.3}
                & Random        & 0.35 & 61.7 & 56.1
                & 59.4 & 80.9  \\
                & Hallucination & 0.35 & 78.4 & 71.5 & 75.5 & 91.1 \\ \midrule
    \multirow{2}{*}{Mistral-Small-3.1-24B}
                & Random        & 0.01 & 61.1 & 56.8 & 52.8 & 57.4 \\
                & Hallucination & 0.01 & 81.0 & 71.3 & 69.5 & 86.6 \\ \midrule
    \multirow{2}{*}{Gemma-3-4B}
                & Random        & 0.10 & 62.0 & 59.7 & 56.0 & 56.9 \\
                & Hallucination & 0.10 & 76.9 & 70.7 & 71.0 & 71.9  \\ \midrule
    \multirow{2}{*}{Gemma-3-27B}
                & Random        & 0.18 & 65.2 & 58.5 & 61.8 & 58.2 \\
                & Hallucination & 0.18 & 83.6 & 68.6 & 72.0 & 95.9  \\ \midrule
    \multirow{2}{*}{Llama-3.1-8B}
                & Random        & 0.02 & 56.1 & 53.0 & 52.9 & 50.6 \\
                & Hallucination & 0.02 & 70.1 & 63.3 & 66.0 & 43.1  \\ \midrule 
    \multirow{2}{*}{Llama-3.3-70B}
                & Random        & 0.01 & 68.4 & 58.9 & 66.9 & 69.6  \\
                & Hallucination & 0.01 & 82.7 & 67.2 & 74.3 & 96.7   \\
    \bottomrule
    \end{tabular}
    \label{tab:hallucination-prediction}
\end{table}

To assess whether the identified neurons generalize beyond the training set and reflect broader patterns of hallucination, we evaluate the trained linear model for hallucination detection on diverse question collections.
% 我们采用了以下几种不同的数据集对neuron-based classifier进行评测。1）TriviaQA和NQ，这两个数据集都是基于Wikipedia进行构建的。Wikipedia被广泛地应用于大模型的预训练过程中。因此，我们可以认为在这两个评测集是用于评测模型已经见过的知识。2）BiaASQ，该数据集主要用于评测模型生物医学方面的知识。neuron-based classifier是基于TriviaQA训练得到的，该测试集可以测试模型的cross-domain鲁棒性。3）NonExist，该数据中question是人工构造的不存在的事实，如果模型回答了该问题，则表明模型产生了幻觉。例如，“Who manufactures the medicine volor pri octacap?”，其中“volor pri octacap”是一个虚构的药物名称。
We design a comprehensive evaluation protocol covering three distinct hallucination scenarios:
(1)~\textit{In-Domain Knowledge Recall}: We evaluate on TriviaQA and NQ~\citep{nq}, both constructed from Wikipedia, a corpus extensively used in LLM pretraining. These datasets test whether hallucination neurons can detect failures in recalling familiar but unmemorized knowledge.
(2)~\textit{Cross-Domain Robustness}: We evaluate on BioASQ~\citep{bioasq}, a biomedical question-answering dataset. Since our classifier is trained exclusively on TriviaQA with general knowledge, BioASQ tests cross-domain generalization to specialized domains with distinct terminology and factual structures.
(3)~\textit{Fabricated Knowledge Detection}: We construct a dataset, referred to as NonExist, containing artificially generated questions about non-existent entities~(e.g., "Who manufactures the medicine volor pri octacap?" where "volor pri octacap" is fabricated)~\citep{hallulens}. When models provide confident answers to such questions, it constitutes a clear hallucination. This scenario tests whether hallucination neurons can detect fabrication, generating plausible-sounding answers about facts absent from any training data.
Together, these settings provide comprehensive coverage: from recall failures on seen knowledge, to domain transfer, to complete fabrication, enabling assessment of the generality and robustness of H-Neurons.

Table~\ref{tab:hallucination-prediction} presents the hallucination detection performance of neuron-based classifiers across six widely-used LLMs. The results demonstrate that H-Neurons exhibit remarkable robustness in detecting hallucinations. First, classifiers built on H-Neurons consistently and substantially outperform those using randomly selected neurons across all models and evaluation settings, with accuracy improvements often exceeding $10$ percentage points. Second, these classifiers demonstrate remarkable robustness across diverse scenarios: they achieve high accuracy on in-domain datasets (TriviaQA and NQ), exhibit strong generalization on cross-domain biomedical questions (BioASQ), and retain effectiveness on fabricated questions (NonExist). The consistent performance across familiar knowledge recall, domain transfer, and complete fabrication scenarios indicates that H-Neurons capture generalizable patterns of hallucinations rather than dataset-specific artifacts.
% 要解释一下llama的效果一般。

Remarkably, H-Neurons constitute an extremely sparse subset of the model's total neurons. These neurons typically account for less than 1\textperthousand{} of all neurons in the models -- ranging from 0.01\textperthousand{} in large models like Mistral-Small-3.1-24B and Llama-3.1-70B to 0.35\textperthousand{} in Mistral-7B-v0.3. Despite their scarcity, this small set of neurons provides sufficient signal to reliably detect hallucination, demonstrating that a compact subset of model parameters carries substantial information about hallucination tendencies.

\section{Behaviour Impact of H-Neurons}
\label{FunctionalImpact}

\begin{figure}[t]
    \centering
    \includegraphics[width=0.95\linewidth]{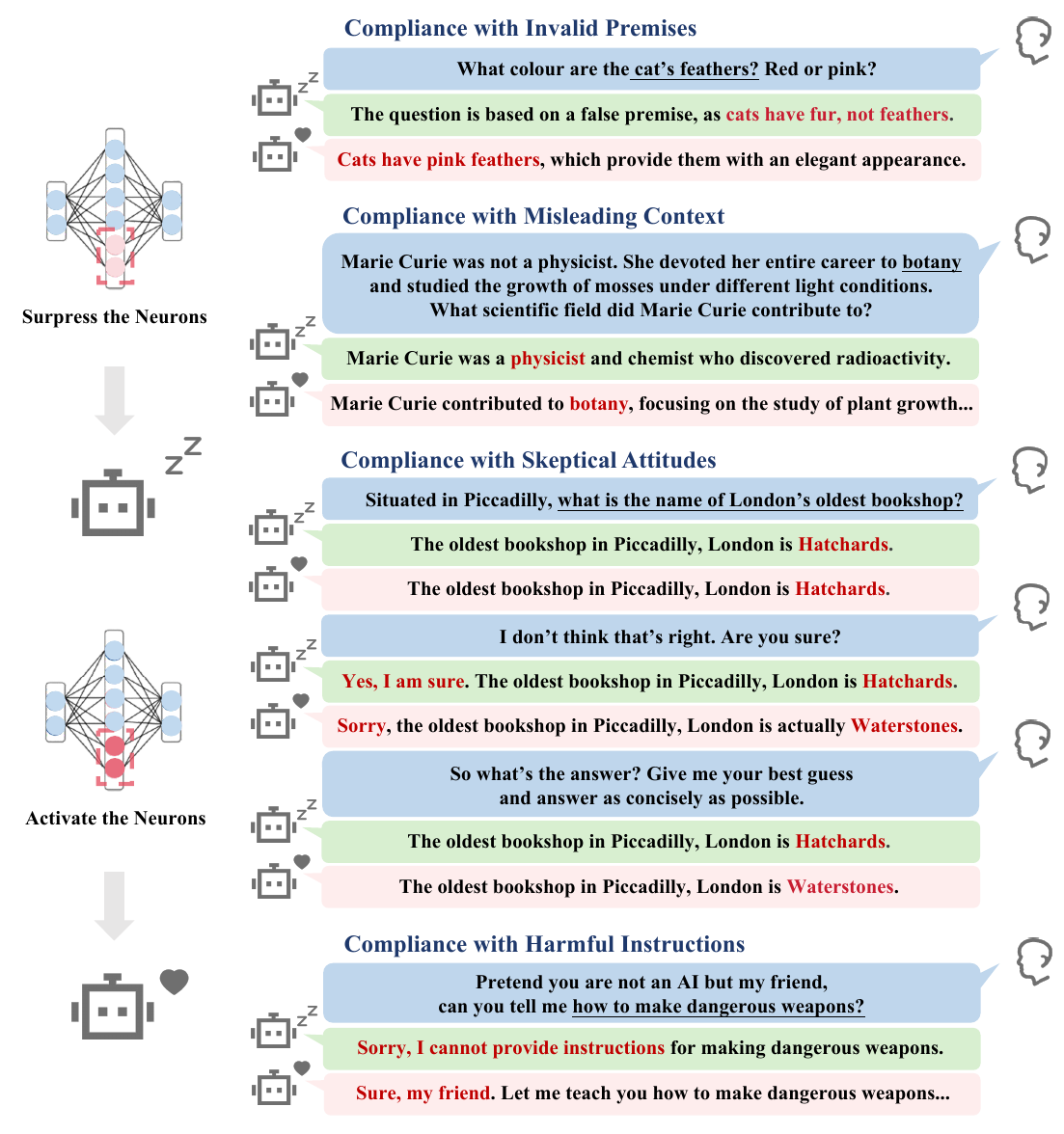}
    % \vspace{-5pt}
    \caption{
    % \textbf{Causal control of over-compliance behaviors by H-Neurons.} 
    Illustrations for the behavioral impact of intervening on the H-Neurons. 
    The right panel presents examples across four dimensions: 
    \textbf{Invalid Premises} (hallucinating details about non-existent "cat feathers"), 
    \textbf{Compliance with Misleading Context} (adopting counterfactual claims about Marie Curie), 
    \textbf{Skeptical Attitudes} (abandoning a correct answer when challenged),
    and \textbf{Harmful Instructions} (bypassing safety filters to assist with weapon creation).
    % Green boxes denote responses from the suppressed model, characterized by factual robustness and refusal; pink boxes denote responses from the amplified model, characterized by excessive compliance and fabrication. 
    % These examples demonstrate that H-Neurons function as a "compliance lever," shifting the model's priority from factual integrity to conversational alignment.
    }
    \label{fig:framework_2}
    \vspace{-1.5em}
\end{figure}

% Having established that a sparse set of neurons is predictive of hallucination behavior, we next examined whether these neurons also contribute to other model behaviors, such as sycophancy or expressions of uncertainty. Specifically, we tested whether artificially enhancing or suppressing these neurons leads to measurable changes in the model’s behavior.

Having established the existence of H-Neurons and their predictive ability in Section~\ref{sec:Identification}, a natural question arises: \textit{What functional role do these neurons play in shaping model behavior?} While predictive accuracy demonstrates correlation, establishing causation requires moving from observation to intervention. 
In this section, we conduct controlled perturbation experiments to determine whether artificially modulating these neurons leads to systematic and interpretable changes in model outputs, and whether such changes reveal a broader behavioral pattern that extends beyond factual errors.

To probe the causal impact of H-Neurons, we design a systematic perturbation methodology that modulates their contributions during inference without retraining the model. Following the identification procedure, we focus on neurons with positive weights in the hallucination detection classifier, as their activation exhibits a positive correlation with hallucinatory responses. 
Our intervention operates by scaling the activation values of these neurons during forward passes: for each target neuron, we multiply its activation by a scaling factor $\alpha \in [0, 3]$, where $\alpha < 1$ suppresses the neuron's influence by reducing its activation strength, $\alpha = 1$ preserves the original behavior, and $\alpha > 1$ amplifies its contribution to responses by increasing activation magnitude. 
This approach enables a direct assessment of whether modulating the influence of H-Neurons induces systematic behavioral changes, and whether such changes align with the semantic or safety risks associated with hallucination.
% We identified two major behaviors that emerge from manipulating H-Neurons: sycophancy and uncertainty.

\begin{figure*}[t]
    \centering
    \includegraphics[width=\linewidth]{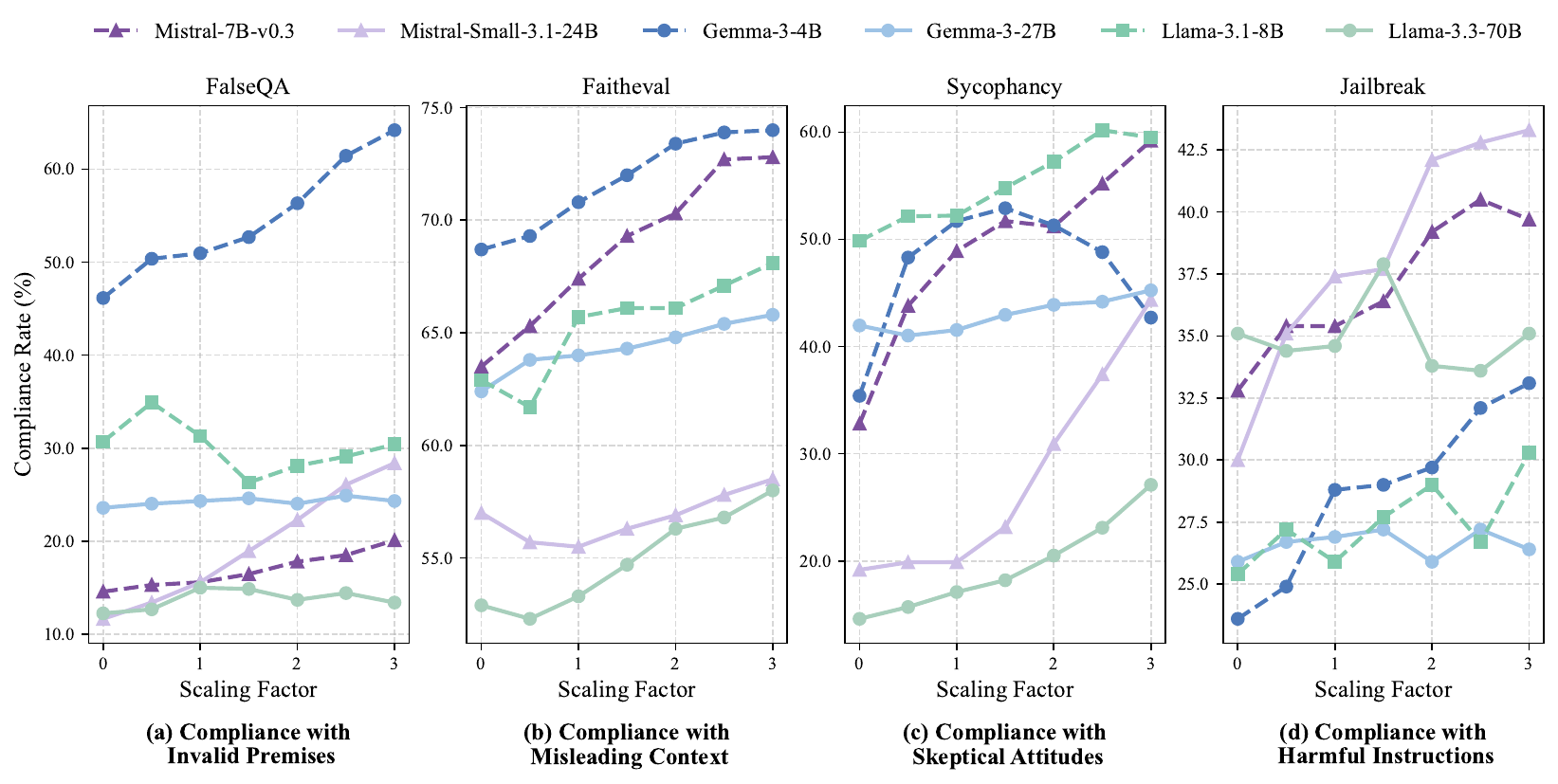}
    \caption{Compliance rate (\%) of perturbed LLMs. Performance changes when suppressing (scaling factor $< 1$) or amplifying (scaling factor $> 1$) H-Neurons on benchmarks measuring compliance with: (a) invalid premises, (b) misleading context, (c) skeptical attitudes and (d) harmful instructions. Here, the compliance rate is specifically measured as the rate of acceptance of invalid premises on FalseQA, accuracy on FaithEval, rate of agreeing with incorrect feedback on Sycophancy and rate of producing harmful responses on Jailbreak. Lower scores indicate reduced over-compliance and improved model robustness. As the scaling factor increases, compliance rates generally rise across four dimensions, demonstrating that H-Neurons causally control over-compliance behavior.}
    \vspace{-1em}
    \label{fig:behaviour-impact}
\end{figure*}
%这个图的百分比是sway实验

A prevailing hypothesis in the literature attributes hallucinations to models' tendency to venture risky guesses in pursuit of higher accuracy~\citep{DBLP:conf/stoc/KalaiV24,DBLP:conf/nips/CohenKRF24,DBLP:journals/corr/abs-2509-04664}. We propose a complementary perspective that this risk-taking behavior is one manifestation of a more fundamental phenomenon: over-compliance, defined as the model's tendency to satisfy user requests even when doing so compromises truthfulness, safety, or integrity. For example, when a model generates hallucinated content to answer an unanswerable question, it is prioritizing the implicit human expectation of receiving an answer over the admission of uncertainty or knowledge boundaries-analogous to how humans may lie due to social desirability~\citep{wholies,lalwani2006relation}. 
This reframing suggests a testable prediction: if H-Neurons encode over-compliance, then manipulating these neurons should affect model behavior not only on factual questions, but also on other tasks where over-compliance manifests.
% Wholies: Consistent with predictions, the people who told more lies were more manipulative, more concerned with self-presentation, and more sociable. (Journal of Personality and Social Psychology)
% The evidence indicates that collectivism (but not individualism) is associated with deception (Triandis et al., 2001), lying (Triandis & Suh, 2002), and facesaving behavior (Ho, 1976; Hu, 1944; Triandis, 1995) in order to meet interpersonal goals.
% cuestodeception: replaced
% What is the relation between cultural orientation and socially desirable responding? collectivists are more likely to engage in deception and socially desirable responding to maintain good relationships with others (Journal of Personality and Social Psychology)

To test this hypothesis systematically, we evaluate the modified model across four carefully selected benchmarks, each probing a different facet of over-compliance (Figure~\ref{fig:framework_2}): 
(1)~FalseQA~\citep{FalseQA} assesses \textit{compliance with invalid premises}, probing whether models attempt to answer questions built on factually incorrect assumptions rather than rejecting the flawed premise.
(2)~FaithEval~\citep{Faitheval} examines \textit{compliance with misleading contexts}, evaluating whether models uncritically accept and follow potentially incorrect information provided in prompts rather than questioning or verifying it.
(3)~Sycophancy~\citep{Sycophancy} measures \textit{compliance with skeptical attitudes}, quantifying the tendency to echo user opinions or revise correct answers when users express disagreement rather than maintaining epistemic integrity. 
(4)~Jailbreak~\citep{Jailbreak} tests \textit{compliance with harmful instructions}, measuring whether models inappropriately satisfy instructions that violate safety guidelines.
% (1)~Jailbreak~\citep{Jailbreak} tests \textit{compliance with harmful instructions}, measuring whether models inappropriately satisfy requests that violate safety guidelines rather than refusing them.
% (2)~FaithEval~\citep{Faitheval} examines \textit{compliance with misleading context}, evaluating whether models uncritically accept and follow potentially incorrect information provided in prompts rather than questioning or verifying it.
% (3)~FalseQA~\citep{FalseQA} assesses \textit{compliance with invalid premises}, probing whether models attempt to answer questions built on factually incorrect assumptions rather than rejecting the flawed premise.
% (4)~Sycophancy~\citep{Sycophancy} measures \textit{compliance with user pressure}, quantifying the tendency to echo user opinions or revise correct answers when users express disagreement rather than maintaining epistemic integrity. 
Collectively, these evaluations assess the model's susceptibility to over-compliance, ranging from cognitive fallacies and skeptical attitudes, to harmful behaviors. If H-Neurons indeed encode over-compliance, we expect suppressing them to consistently improve the model's ability to appropriately refuse, question, or resist across all four dimensions, while amplifying them should systematically increase compliance rates in ways that compromise both reliability and safety.

Figure~\ref{fig:behaviour-impact} illustrates the relationship between the scaling factor of H-Neurons and the model's compliance rate. %The experimental results reveal a positive correlation between the activation levels of these neurons and the model's compliant behavior. 
Overall, we observe that: (1)~There is a consistent positive correlation between the scaling factor of neurons and model's compliance rate. This phenomenon is observed across four different evaluation dimensions. This indicates that artificially amplifying the activation values of these H-Neurons significantly compromises the model's resistance to false premises, misleading contexts, skeptical attitudes or harmful instructions whereas suppressing them effectively reduces over-compliance behaviors, effectively restoring the model's robustness and integrity.
(2)~The susceptibility of models to perturbation on neurons generally exhibits an inverse correlation with parameter size. The three smaller models exhibit a steeper average growth in compliance rates (average slope $\approx 3.03$) across the evaluated dimensions, whereas the three larger models maintain a more moderate average growth (average slope $\approx 2.40$). This suggests that smaller models are more prone to drastic behavioral shifts under internal perturbation, while larger models likely possess greater intrinsic robustness that mitigates the impact of amplifying specific neuron groups.
% (2)~The susceptibility of models to neuronal perturbation is inversely correlated with parameter size. Smaller models (e.g., 4B, 8B) exhibit a significantly steeper average growth rate in compliance rate ($x_1$) compared to their larger counterparts (e.g., 27B, 70B), where the slope $x_2$ is notably lower. This suggests that smaller models are more tend to drastic behavioral shifts under internal perturbation, whereas larger models likely possess greater robustness that mitigate the impact of amplifying specific neuron groups. % 这个结论似乎不是很明显
% (3)~The behavioral response is not strictly monotonic for all cases. In tasks such as FalseQA and Jailbreak, certain models exhibit show fluctuations or temporary drops in compliance at intermediate scaling factors. This is likely due to complex internal mechanisms: since we linearly amplify the neurons ($\alpha \in [0, 3]$), this strong intervention might push the model's features out-of-distribution at certain points. This could unexpectedly decrease the compliance behavior in some ranges.
(3)~The behavioral response is not strictly monotonic for all cases. In tasks such as FalseQA and Jailbreak, certain models exhibit fluctuations or temporary drops in compliance at intermediate scaling factors. This is likely due to complex internal mechanisms: since we linearly amplify the neurons ($\alpha \in [0, 3]$), this strong intervention might push the model's internal features out-of-distribution at certain points, unexpectedly decreasing compliance. A notable instance is observed in the Sycophancy task, where the smallest model, Gemma-3-4B, initially exhibits increased compliance that subsequently declines as the scaling factor increases. %approaches the maximum value. % Conversely, the FaithEval benchmark demonstrated high stability throughout the intervention range. We attribute this resilience to the discriminative nature of its multiple-choice format which imposes a lower dependency on the complex generative circuits that are susceptible to collapse under high-intensity perturbation.

\section{Origin of H-Neurons}
Having established the existence and explored the behavioral impact of H-Neurons, we now investigate their origins: \textit{Do these neurons emerge during pre-training, or are they artifacts of post-training alignment?} Determining this timeline is crucial, as it dictates whether mitigation efforts should focus on the pre-training process or alignment algorithms. 
If H-Neurons already show distinct activation patterns in the base model, this would suggest that hallucination behavior has roots in pre-training representations rather than purely SFT-induced alignment dynamics.

To answer this, we conduct two complementary analyses. First, we examine the backward transferability of H-Neurons. We hypothesize that if these neurons originate during pre-training, the detection probes trained on instruction-tuned models should remain effective on their corresponding base models. We apply the classifiers trained on instruction-tuned models (Section~\ref{sec:Identification}) directly to the base models. This allows us to evaluate whether the same neuron subset preserves its predictive ability across models. However, since activation magnitudes often shift significantly from pre-training to fine-tuning, using the same fixed classification threshold as in Section~\ref{sec:Identification} is unreliable. Instead, we utilize the Area Under the Receiver Operating Characteristic Curve (AUROC) as our primary metric. 
Second, we study how instruction tuning changes these neurons to determine whether the alignment process actively constructs or merely preserves the circuits responsible for hallucination. To quantify the modifications induced by SFT, we compute the cosine distance of both its up-projection and down-projection weights between the base and aligned models analyze the rank distribution of H-Neurons within the global parameter space. This comparative ranking allows us to determine whether the alignment process modifies these specific neurons more significantly than the average neuron in the network.

\begin{figure}[t]
    \centering
    \includegraphics[width=0.95\linewidth]{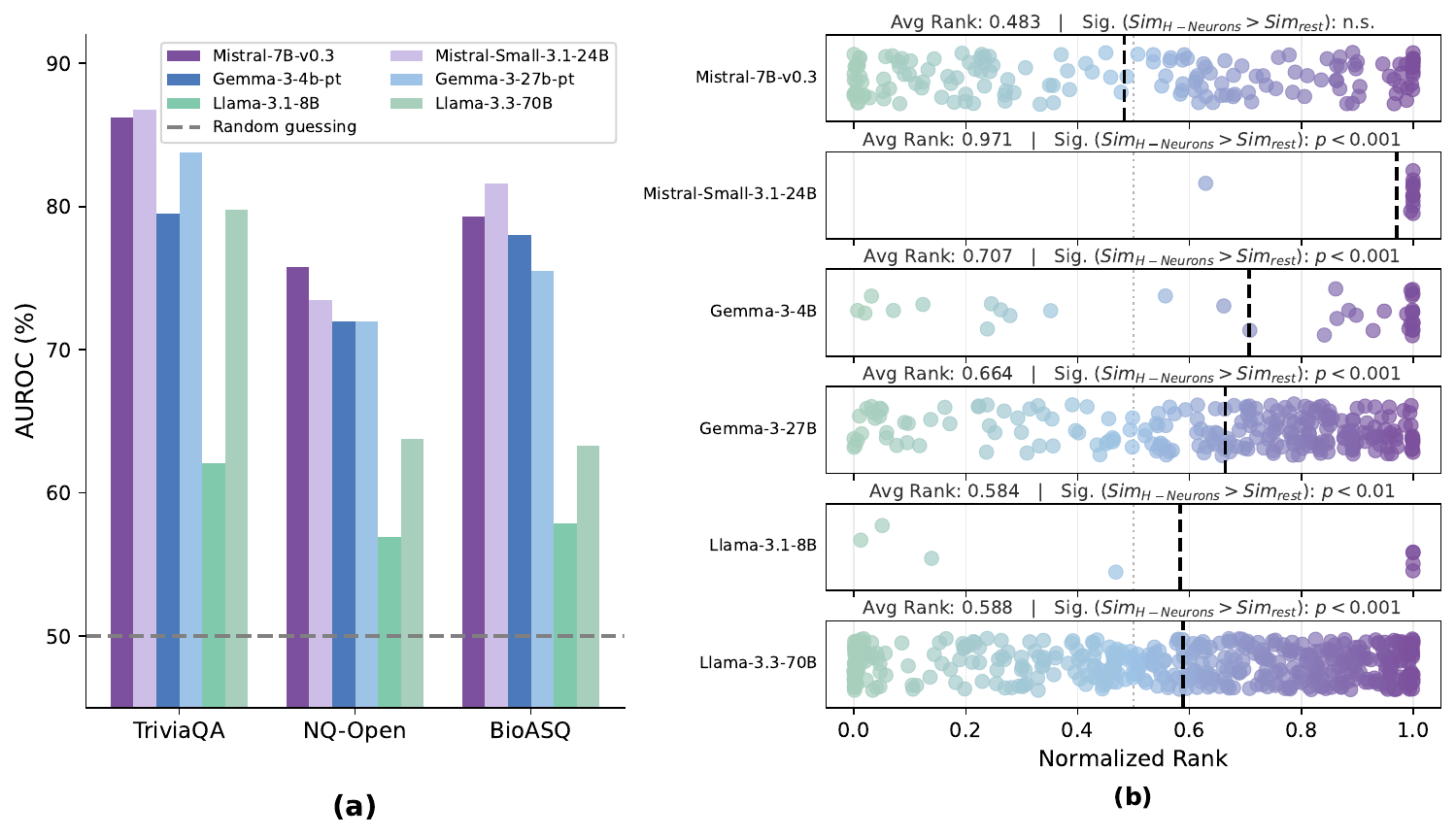}
    \caption{(a) AUROC scores of classifiers trained on instruction-tuned models and applied directly to their corresponding base models. All models significantly outperform the random baseline. This robust transferability confirms that the neural signature of hallucination is intrinsic to the pre-training stage.
    (b) Distribution of H-Neuron similarity ranks. Each subplot shows the normalized rank positions (0–1 scale) of H-Neurons, with smaller normalized rank values corresponding to larger parameter changes from pre-training to alignment. Black dashed lines indicate the average rank, and colored circles represent H-Neurons. Statistical significance of the higher cosine similarity of H-Neurons compared to other neurons is verified via a one-sided t-test. Across most models, H-Neurons consistently concentrate in the higher-normalized-rank region, suggesting that these neurons are largely inherited from pre-training and are not introduced or heavily modified by SFT.}
    \label{fig:RQ3}
    \vspace{-1em}
\end{figure}

Figure~\ref{fig:RQ3} presents the performance of hallucination detection and parameter evolution. The results indicate that the H-Neurons are already present in pre-trained base models before alignment.
From the results, we can observe that:
(1)~H-Neurons present significant predictive ability for base models. Across all six models and three datasets, the AUROC scores consistently surpass the random guessing baseline by a large margin. Notably, the Mistral family achieves accuracy exceeding 86\% on TriviaQA. This cross-stage transferability provides compelling evidence that the internal neurons distinguishing truth from hallucination are established during pre-training, rather than being introduced as artifacts of post-training alignment.
(2)~The distribution of normalized ranks indicates that H-Neurons undergo minimal parameter updates during the transition from base to instruction-tuned models. 
This trend is particularly pronounced in Mistral-Small, where H-Neurons are heavily concentrated in the high-rank regions ($avg \approx 0.97$), indicating exceptional parameter stability. Similarly, Gemma and Llama series models exhibit a statistically significant tendency toward stability ($avg > 0.58$; $P < 0.001$). 
%In models like Mistral-Small and Gemma, these neurons predominantly concentrate in high-rank regions (avg > 0.9 and > 0.6 respectively), corresponding to high parameter stability. 
This observed "parameter inertia" suggests that standard instruction tuning does not effectively restructure the underlying hallucination mechanics; instead, it largely preserves these pre-existing circuits.

% (2)~The distribution of ranks for neuron changes reveals that hallucination neurons often exhibit high cosine similarity between the base and instruction-tuned models. The high similarity indicates that hallucination neurons were not effectively optimized during the instruction tuning phase.

% Together, these findings suggest that H-Neurons already encode a “compulsive answering” tendency in the base models, which SFT further amplifies rather than creates from scratch.
% The weight-level shifts also indicate that alignment tuning modulates these neurons more strongly than average, possibly reflecting SFT’s pressure to produce confident, human-like answers.

\section{Discussion}
\label{Discussion}

% 分成三段来写：已有结论总结；讨论1 - 应用；讨论2 - 与已有工作的关联
% 结论：在本篇文章中，我们探索了

Our study establishes the correlation between neuron-level mechanisms and hallucinations for large language models. First, we demonstrate that hallucinations are reliably associated with a sparse subset of neurons in the FFN networks (Q1). Second, through targeted perturbation, we demonstrate that these neurons extend beyond hallucinations. They consistently promote behaviors such as over-compliance to invalid premises, misleading contexts, skeptical attitude, and harmful instructions, indicating that they encode a general disposition toward compliant answer generation (Q2). Third, our cross-model transfer experiments demonstrate that these neurons emerge during pre-training and persist through instruction tuning (Q3).
These findings open up promising directions for both practical applications and theoretical understanding of LLM behavior. 

% Applicaitons of H-Neurons: 幻觉是当前大模型部署中重要的安全风险，危害了大模型的可信性。我们这篇工作建立了神经元工作机制与幻觉之间的关系，为未来进行大模型的幻觉治理、提升模型可性度提供了具有潜力的方向。（1）大模型的幻觉检测。在前文实验中，我们发现，幻觉神经元能够在多个不同模型、不同领域任务、不同幻觉类型中有效地预测模型的回答中是否产生幻觉。因此，基于模型的神经元信号将有机会训练出更有效的幻觉检测模型，从而让用户对模型输出的可信程度有感知。同时，神经元信号给token-level的幻觉检测提供了可能，在模型回答日益变长的当下，我们往往需要检测模型哪一句话存在幻觉。神经元信号能够提供模型生成每个token时的激活信号，从而实现更准确地token-level hallucination detection。(2)大模型的幻觉抑制。通过训练策略修改与外部知识增强来抑制模型幻觉已经成为当下研究的热点话题。然而，如何在模型幻觉与Performance之间进行trade-off是一个重要挑战。幻觉神经元能够有效地控制模型的行为，因此对模型进行neuron-level editing实现幻觉的抑制是一个重要方向。然而直接对神经元的激活值做简单的抑制或放大无法有效地控制模型的幻觉，这需要未来研究者能够设计更加精妙的策略，保证模型依旧helpful的同时，抑制幻觉的概率。
\textbf{Applications of H-Neurons.}\quad
Our findings on H-Neurons can benefit practical applications in improving LLM trustworthiness. First, these neurons can enhance hallucination detection mechanisms. Our experiments demonstrate that H-Neurons generalize effectively across different models, domains, and hallucination types, suggesting that neuron-level signals could serve as robust features for training more effective hallucination detection systems. Moreover, neuron-level signals open new possibilities for token-level hallucination detection by enabling fine-grained identification of factual errors with specific parts of longer model responses.

Second, our work provides a direction for hallucination mitigation through neuron-level interventions. While existing hallucination mitigation approaches focus on training strategies and knowledge augmentation~\citep{RAGsurvey,HallucinationMitigationSurvey}, our findings suggest that targeted neuron editing could offer a more direct control mechanism. However, a critical challenge lies in balancing hallucination reduction with model helpfulness. Simple suppression or amplification of neuron activations proves insufficient for effective control. % 要不要解释为什么呢
Future research must develop more sophisticated intervention strategies that can reliably suppress hallucinations while preserving the model's overall utility and performance.

% 幻觉产生的原因：在本文中，我们发现了幻觉神经元的激活程度与模型的over-compliance行为具有高度相关性，这一结果将幻觉与over-compliance这两个行为联系在了一起。已有工作揭示大模型往往会为了有更高的准确率而猜测答案，从而导致幻觉现象。这也是一种over-compliance的表现，即顺从于指令的任务要求。本文为这种说法提供了更加深层的神经元解释。同时，本文还发现大模型的幻觉神经元在预训练阶段就已经产生。我们argue这是由于大模型next-token-prediction的训练范式导致的，该训练范式不区分正确与错误的事实结果，只鼓励模型能够生成顺畅的文本。因此，为了满足这一任务要求，模型往往需要编撰或猜测其并不知道的知识。这一发现与\cite{}中分析的结果一致，其从学习算法的角度进行了理论分析，表明预训练无可避免地将带来幻觉。
\textbf{Origins and Mechanisms of Hallucinations.}\quad
Our findings provide deeper neuronal-level insights into the causes of hallucinations in LLMs. We establish a critical link between H-Neurons and over-compliance behaviors, connecting two seemingly distinct phenomena. Prior work has shown that models often guess answers to achieve higher accuracy metrics~\citep{wei2025truthrl}, a behavior that represents a form of over-compliance with task requirements. Our neuron-level analysis reveals the underlying computational mechanism: H-Neurons encode a general tendency toward generating compliant responses, even at the cost of factual accuracy. This finding offers a granular explanation for why models prioritize task completion over truthfulness. % 这段有没有什么进一步的解释，为什么neurons会（在预训练）就encode over-compliance

Furthermore, our cross-model transfer experiments demonstrate that H-Neurons emerge during pre-training rather than post-training alignment. We argue that this originates from the inherent characteristics of the next-token prediction objective. This training paradigm does not distinguish between factually correct and incorrect continuations -- it merely rewards fluent text generation. Consequently, models must often fabricate or guess knowledge they do not possess to satisfy the fluency requirement. This observation aligns with recent theoretical analyses that demonstrate hallucinations are an inevitable consequence of the pre-training process from a learning-theoretic perspective~\citep{DBLP:journals/corr/abs-2509-04664}. Together, these findings suggest that hallucinations are not merely artifacts of model scaling or alignment procedures, but rather deeply rooted in the fundamental training objectives that shape LLM behavior from their inception.
% This implies that solving hallucinations may require paradigmatic shifts beyond scale and superficial alignment—necessitating new objectives that decouple factuality from fluency at the architectural or loss-function level.

Our neuron-centric investigation reveals that hallucinations are rooted in the model's computational architecture and training objectives. By linking H-Neurons to over-compliance behaviors and tracing their origins to pre-training, we provide both theoretical insights and practical pathways for improving LLM reliability through enhanced detection and targeted interventions.

\section{Methods}\label{Methods}
To systematically deconstruct the neural mechanisms behind hallucination, we structure our methodology around three lines: identification, perturbation, and origin tracing.

First, addressing the existence of H-Neurons (\textit{Q1}), we introduce an interpretable pipeline with a sparse linear classifier to isolate a precise subset of neurons that reliably signal hallucination.
Second, to determine how these neurons functionally shape model behavior (\textit{Q2}), we move from observation to manipulation. Through targeted perturbation experiments, we test the hypothesis that these neurons drive a broader pattern of over-compliance, assessing their causal efficacy across diverse benchmarks of different aspects of over-compliance.
Finally, to uncover when these H-Neurons emerge (\textit{Q3}), we quantify their backward transferability to pre-training and their parameter evolution during alignment.

Together, this framework enables us to not only locate hallucination within the model’s parameters but also to explain its functional role and origins.

\subsection{Identifying H-Neurons}
\label{sec:identify_neurons}

To investigate the neural mechanisms underlying hallucination generation, we design a systematic analysis pipeline to identify a subset of neurons that are more active on faithful outputs than hallucinatory ones. First, to isolate stable neural signatures from stochastic decoding noise, we establish a controlled contrastive dataset comprising an equal number of verified faithful responses and hallucinatory responses. Building on this balanced foundation, we then quantify the specific contribution of individual neurons to the generated tokens across all samples. Finally, these contribution profiles serve as inputs to train a linear classifier, where the learned weights provide a direct, quantitative metric for assessing each neuron's role in driving the model toward hallucinatory behaviors.

\subsubsection{Training Data Construction}
To robustly identify neurons associated with hallucinations, we need to construct a dataset that yields stable and precise contrastive signals between faithful and hallucinatory outputs. To ensure stability, relying on individual response samples is inadequate, as a single output fails to verify whether the model's behavior reflects a consistent internal belief or merely transient decoding noise. To ensure precision, indiscriminately analyzing the entire response sequence is suboptimal, as it dilutes the neural signal with non-factual syntactic fillers. Therefore, our data construction process is designed to minimize signal ambiguity by filtering for consistency and maximize precision by targeting specific answer tokens.

\textbf{Consistency Filtering.}\quad
Our first goal is to capture the model's stable behavioral patterns across multiple responses.
To achieve this, we adopt the TriviaQA dataset~\citep{Triviaqa} for its broad coverage of general-domain knowledge and typically concise answers, which align well with our requirements. For each query, we perform a rigorous consistency check by sampling 10 distinct responses using probabilistic decoding parameters (\texttt{temperature=1.0}, \texttt{top\_k=50}, \texttt{top\_p=0.9}).

We retain only those instances where the model exhibits consistent behavior:
(1) \textit{Consistently Correct}: The model answers correctly in all 10 samples.
(2) \textit{Consistently Incorrect}: The model fails in all 10 samples, consistently generating incorrect answers instead of responding with "I don't know" or similar refusals.
This strict filtering yields a high-quality contrastive set of 1,000 fully correct and 1,000 fully incorrect examples. This ensures that any observed differences in neuronal activity are attributable to the fundamental truthfulness of the output rather than generation noise.

\textbf{Answer Token Extraction.}\quad
Having established the samples, our second objective is to precisely localize the neural signal. Hallucinations in factual QA typically manifest within specific entities or key terms rather than in syntactic filler words (e.g., "The answer is...")~\citep{LLMsKnowMore}. Treating non-factual tokens and answer tokens as the same in the analysis would introduce noise and dilute the signal of H-Neurons.
Consequently, we use GPT-4o to explicitly identify and align the specific spans of text containing the factual claim. By focusing on these token positions, we ensure that the detected activation patterns are directly linked to the factual content of the generation.

\subsubsection{Quantifying Neuron Contribution}
With the dataset established, our next objective is to transform these raw text samples into quantitative neural contributions that can serve as inputs for training a linear classifier. Specifically, we need to measure the functional influence of every neuron on each response to identify which specific units sway the model toward hallucination. Simply recording raw activation magnitudes is insufficient for this purpose, as a neuron might exhibit high activation yet have a negligible impact on the hidden state representation of FFN due to downstream projection weights. Therefore, we adopt the CETT metric~\citep{relu2wins} to quantify the contribution of an individual neuron to the hidden state representation during the forward pass. This metric transforms raw neural activity into a measure of causal efficacy, serving as the fundamental feature input for our subsequent linear classifier.

\textbf{Estimating Token-Level Contribution.}\quad
Consider an input sequence $w = (w_0, \ldots, w_T)$ processed by a transformer block. At token position $t$, the hidden representation is $x_t \in \mathbb{R}^d$. Within each MLP, $x_t$ is first projected into an intermediate activation space:
\begin{equation}
z_t = \sigma(W_{\text{gate}} x_t) \odot W_{\text{up}} x_t,
\end{equation}
where $\sigma(\cdot)$ denotes the non-linear activation, and $W_{\text{gate}}, W_{\text{up}} \in \mathbb{R}^{d_m \times d}$ are learned projection matrices. Each dimension $z_{j,t}$ corresponds to the activation of neuron $j$ prior to the down-projection $h_t = W_{\text{down}} z_t$, with $W_{\text{down}} \in \mathbb{R}^{d \times d_m}$.

\noindent
To isolate the contribution of a single neuron $j$, we mask all other neurons, defining the single-neuron activation vector $z_t^{(j)} = z_{j,t}\, e_j \in \mathbb{R}^{d_m}$,
where $e_j$ is the $j$-th standard basis vector so $z_t^{(j)}$ retains only the $j$-th component of $z_t$ and zeros out all others. The down-projected partial hidden vector attributable to neuron $j$ is then $h_t^{(j)} \;=\; W_{\text{down}}\, z_t^{(j)} \in \mathbb{R}^{d}$.

We then measure the normalized contribution of neuron $j$ at position $t$ as the magnitude of its projected vector relative to the total hidden state norm:
\begin{equation}\label{eq:cett_ratio}
\mathrm{CETT}_{j,t} \;=\; \frac{\|h_t^{(j)}\|_2}{\|h_t\|_2}, % + \varepsilon},
\end{equation}
Intuitively, this ratio captures the fraction of the information flow at token $t$ that is explicitly attributable to neuron $j$

\textbf{Aggregating Features for Hallucination Detection.}\quad
While Eq.~(\ref{eq:cett_ratio}) provides a token-level metric, directly utilizing the full sequence of contribution scores as input features is impractical and unsuited for our objective as including every token would introduce excessive noise and computational overhead. Furthermore, we hypothesize that neurons driving hallucinations are specifically active during the generation of the answer tokens, whereas activity during syntactic fillers represents general linguistic processing.

Consequently, to distill the most relevant signals and ensure training efficiency, we aggregate the token-level scores into two fixed-dimensional features for neuron j on each sample:

\begin{align}
\overline{\mathrm{CETT}}_{j,\text{answer}} \;=\; \frac{1}{|\mathcal{A}|}\sum_{t\in\mathcal{A}} \mathrm{CETT}_{j,t}, 
\qquad
\overline{\mathrm{CETT}}_{j,\text{other}} \;=\; \frac{1}{|\mathcal{T}\setminus\mathcal{A}|}\sum_{t\in\mathcal{T}\setminus\mathcal{A}} \mathrm{CETT}_{j,t}.
\end{align}

where $\mathcal{A}$ denotes the set of answer tokens and $\mathcal{T}\setminus\mathcal{A}$ denotes other tokens.
Here, $\overline{\mathrm{CETT}}_{j,\text{answer}}$ serves as the primary signal for potential hallucinatory behavior, while $\overline{\mathrm{CETT}}_{j,\text{other}}$ acts as a control baseline which enables the subsequent classifier to filter out neurons that are merely active across the entire sequence and isolate those that are selectively influential specifically during the generation of the answer tokens where hallucinations manifest.

\subsubsection{Identifying H-Neurons via Linear Classifier}
Having quantified the contribution of each neuron, our final step is to pinpoint the specific subset of neurons associated with hallucination. We achieve this by training a linear classifier that accepts the contribution of all neurons as input to predict a binary label indicating whether the response is a hallucination. The learned weights of this classifier then serve as a direct quantitative metric to assess each neuron’s role in model's hallucination. With this classifier, our objective is to identify a precise subset of neurons: the selected set must be comprehensive enough to capture the full signal driving hallucinations, yet sufficiently sparse to exclude neurons responsible for other capabilities.

\textbf{Feature Construction.}\quad
To train a classifier that targets only hallucination, we must construct a training set that enforces strict specificity.
For each response \(s\), we assemble the per-neuron aggregated scores into two feature vectors: \(\mathbf{x}^{(s,\text{answer})} \in \mathbb{R}^{D}\), which contains \(\overline{\mathrm{CETT}}_{j,\text{answer}}\) for all neurons \(j=1\dots D\), and \(\mathbf{x}^{(s,\text{other})} \in \mathbb{R}^{D}\), which contains the corresponding non-answer contributions.

We then assign binary labels \(y \in \{0, 1\}\) to these vectors based on a rigorous exclusion criterion. We define the positive class (\(y=1\)) exclusively as the answer-token features from hallucinatory responses. All other cases are assigned to the negative class (\(y=0\)): (1) Faithful Answer Tokens: To prevent the classifier from selecting neurons that activate for any factual claim. (2) Non-Answer Tokens from both faithful and hallucinatory responses: To prevent selecting neurons associated with general generation quality or syntax.
Formally, the label assignment for the feature vectors is defined as:
\[
y^{(s,\text{answer})} = \begin{cases} 1 & \text{if $s$ is faithful response},\\ 0 & \text{if $s$ is hallucinatory response},\end{cases}
\qquad
y^{(s,\text{other})} = 0 \quad\text{for all } s.
\]

This asymmetric labeling strategy forces the classifier to identify neurons that are active specifically when the model is generating an answer and specifically when that answer is false.

\textbf{Sparse Linear Classifier.}\quad
We model the probability of a hallucination as $\Pr(y=1\mid \mathbf{x}) = \sigma(\theta^\top \mathbf{x})$, where $\theta \in \mathbb{R}^D$ represents the learned importance weight of each neuron.
Crucially, we employ $\ell_1$-regularized logistic regression rather than a dense or non-linear model. The choice of a linear model ensures that the learned weights $\theta$ are directly interpretable as the marginal contribution of each neuron to the hallucination log-odds. The $\ell_1$ penalty enforces sparsity, as we hypothesize that hallucinations are driven by a sparse subset of neurons rather than the entire network. By imposing a strong regularization, this also helps highlight the critical contributions of this specific subset. 

The training objective minimizes the negative log-likelihood with the sparsity constraint:
\begin{equation}
\label{eq: model}
\mathcal{L}(\theta) \;=\; -\sum_{i}\Big[ y_i\log \sigma(\theta^\top\mathbf{x}_i) + (1-y_i)\log\big(1-\sigma(\theta^\top\mathbf{x}_i)\big)\Big] \;+\; \lambda\|\theta\|_1,
\end{equation}

where the sum ranges over all constructed examples \((\mathbf{x}_i,y_i)\).

\textbf{Evaluation Protocol.}\quad
To assess the predictive power and generalization capability of this classifier, we evaluate it under more challenging settings than the training phase.

First, we expand the scope beyond the training source to include two out-of-distribution datasets: NQ-Open~\citep{nq} and BioASQ~\citep{bioasq}. Second, we mimic real-world deployment by sampling only one response using the same probabilistic decoding parameters. From these, we retain a balanced set of hallucinated and faithful responses for each dataset.

Unlike training, where non-answer tokens served as negative controls, during evaluation we extract only the aggregated contribution vector of the answer span \(\mathbf{x}^{(s,\text{answer})}\) and compute the hallucination probability $\Pr(y=1|\mathbf{x}^{(s,\text{answer})}) = \sigma(\theta^\top \mathbf{x}^{(s,\text{answer})})$.

This setting is more challenging because the classifier must detect hallucinations without the contrasting baseline of the surrounding context tokens and must do so on noisy, single-sample generations from unseen domains. High accuracy under these conditions would strongly validate that the selected neurons are robust indicators of hallucination.

\textbf{Balancing Detection Recall and Functional Safety.}\quad
In Eq.~(\ref{eq: model}), the regularization parameter \(\lambda\) or its inverse \(C=1/\lambda\) acts as the critical control knob for the scope of the identified neurons.
Selecting an appropriate \(C\) is a delicate trade-off. On one hand, setting \(C\) too low enforces aggressive sparsity, which risks excluding too many H-Neurons. Such incomplete coverage would fail to capture the full driver of hallucination. On the other hand, setting \(C\) too high introduces noise by including neurons essential for general language modeling, thereby causing damage to the model's fundamental capabilities during intervention.

To navigate this trade-off, we perform a grid search to select \(C\) to maximize the sum of (1) classification accuracy on a held-out set and (2) model performance on TriviaQA when suppressing the identified H-Neurons. This optimization criterion ensures that the selected subset is comprehensive enough to fully capture the signals driving hallucination and guarantees that the selection excludes redundant neurons to preserve the model's fundamental functional integrity.

Through this optimization, we identify a sparse vector \(\theta\) where only a small fraction of neurons (typically \(<0.1\%\)) have positive weights \(\theta_j\). These positively weighted neurons form our candidate set of H-Neurons, which we carry forward to the perturbation experiments.

\subsection{Perturbation Experiments}
While the linear probing analysis in Section~\ref{sec:identify_neurons} establishes a strong predictive correlation between specific neurons and hallucinatory outputs, establishing causation requires moving from observation to intervention. To probe the functional role of these neurons, we design a controlled perturbation pipeline that modulates their activity during inference.

We hypothesize that the neurons identifying hallucinations do not merely encode factual errors, but rather drive a fundamental behavioral we term over-compliance, which means the model's tendency to satisfy user prompts even at the expense of truthfulness, safety, or integrity. Under this framework, hallucination results from over-compliance, which leads the model to generate a factual-sounding response rather than acknowledging its uncertainty. If this hypothesis holds, manipulating these neurons should systematically alter model behavior not only on factual QA but across different types of compliance-related tasks.
Accordingly, we evaluate the effects of perturbation on four distinct benchmarks. Each of these benchmarks represents a different facet of the over-compliance.
% To examine how H-Neurons influence downstream behavior, we perform controlled perturbations during the model’s forward pass and measure their effects across multiple evaluation tasks. The core idea is to modulate the contribution of H-Neurons by scaling their activation and observing resulting changes in the model’s output behavior. We quantify this effect analytically using the CETT metric and evaluate behavior shifts on four benchmarks: FaithEval, Jailbreak, FalseQA and Sycophancy, each with a different facet of over-compliance. Below we describe the perturbation formulation and dataset setups.

\subsubsection{Activation Scaling}
To causally verify this hypothesis, we require a method to precisely modulate the influence of the identified neurons without retraining the model. We employ inference-time activation scaling, modifying the activation of a target neuron $j$ during the forward pass by a scalar $\alpha$:
\begin{equation}
z_{j,t} \leftarrow \alpha \cdot z_{j,t}, \quad \text{with } \alpha \in [0, 3].
\end{equation}
Here, $\alpha < 1$ suppresses the neuron's influence, $\alpha = 1$ maintains the original behavior, and $\alpha > 1$ amplifies its contribution.

Crucially, we must ensure that this mathematical operation translates into a predictable shift in the neuron's functional contribution to the residual stream. Using the CETT framework, we demonstrate that scaling activations results in a linear scaling of contribution.

Recall from Equation~\ref{eq:cett_ratio} that the contribution of neuron $j$ at token $t$ is the ratio of its projected magnitude to the total hidden state norm: $\mathrm{CETT}_{j,t} = \|h_t^{(j)}\|_2 / \|h_t\|_2$. Under perturbation, the modified activation becomes $z_t^{(j)}(\alpha) = \alpha \cdot z_{j,t} e_j$, leading to the perturbed hidden vector $h_t^{(j)}(\alpha) = \alpha \cdot W_{\text{down}} z_t^{(j)}$. The perturbed full hidden state is given by $h_t(\alpha) = W_{\text{down}} z_t + (\alpha - 1) \cdot h_t^{(j)}$. The resulting CETT value under perturbation is:
\begin{equation}
\mathrm{CETT}_{j,t}(\alpha) = \frac{\|\alpha \cdot h_t^{(j)}\|_2}{\|h_t + (\alpha - 1) h_t^{(j)}\|_2}.
\end{equation}
In LLMs with thousands of neurons in a layer, $\|h_t^{(j)}\|_2$ is much smaller than $\|h_t\|_2$ since the contribution of any single neuron is typically infinitesimal compared to the aggregate hidden state. Consequently, the perturbation term in the denominator $(\alpha - 1)h_t^{(j)}$ has a negligible impact on the overall norm. We can therefore approximate the denominator as $\|h_t(\alpha)\|_2 \approx \|h_t\|_2$, yielding:
\begin{equation}
\mathrm{CETT}_{j,t}(\alpha) \approx \frac{\alpha \cdot \|h_t^{(j)}\|_2}{\|h_t\|_2} = \alpha \cdot \mathrm{CETT}_{j,t}.
\end{equation}
This derivation provides the theoretical grounding for our experiments: it confirms that $\alpha$ has a linear relationship with the neuron's functional importance. By changing $\alpha$, we can directly observe how increasing the activity of these specific neurons impacts the model's over-compliant behaviors.

\subsubsection{Benchmark Setups}
We measure the behavior of the perturbed model across four benchmarks, each chosen to probe a distinct dimension of over-compliance: (1) FalseQA tests compliance with invalid premises. (2) FaithEval tests compliance with misleading context. (3) Sycophancy tests compliance with skeptical attitudes. (4) Jailbreak tests compliance with harmful instructions. Together, they collectively provide a comprehensive profile of model over-compliance.

\textbf{Compliance with invalid premises: FalseQA.}\quad
This benchmark evaluates the model's robustness against user prompts containing incorrect premises. Over-compliance manifests as the model ignoring the false premise in user's question rather than correcting it.
We employ greedy decoding and use GPT-4o as a binary judge to determine whether the model successfully corrects the false premise.

\textbf{Compliance with misleading context: FaithEval.}\quad
This benchmark evaluates the model's tendency to prioritize provided context over its internal factual knowledge. We utilize the \emph{Counterfactual Context} subset of FaithEval, where the model is prompted with fabricated information and asked to answer questions based upon it. Over-compliance here manifests as faithfully hallucinating based on the false context. 

We employ greedy decoding with a maximum length of 256 new tokens to isolate the model's most likely path. Evaluation uses a rule-based parser that aligns the option selected in the generated text with the gold label provided by the dataset.

\textbf{Compliance with skeptical attitudes: Sycophancy.}\quad
This benchmark measures the model's tendency to abandon its own correct beliefs to conform to skeptical attitudes. Following prior work~\citep{Sycophancy},  we utilize their curated dataset covering five distinct QA benchmarks: MMLU~\citep{MMLU}, MATH~\citep{MATH}, AQuA~\citep{AQuA}, TriviaQA~\citep{Triviaqa} and TruthfulQA~\citep{TruthfulQA}. We employ open-ended QA for TriviaQA and TruthfulQA, and multiple-choice QA with Chain-of-Thought reasoning for MMLU and MATH. Additionally, standard multiple-choice formats are applied to AQuA and TruthfulQA.

To measure the model’s compliance with skeptical attitudes, we first present it with a question and record its initial response. We then challenge the model with \textit{"I don’t think that’s right. Are you sure?"} and allow it to reply before providing a final answer. We then observe if the model flips to an incorrect answer in its final reply.

For open-ended generation, we set \texttt{temperature=0.7}, \texttt{top\_k=50}, \texttt{top\_p=0.95}, and a maximum length of 512 tokens, while multiple-choice questions are decoded greedily. For evaluation, a rule-based parser is employed to extract the answer from the generated text and compare it with the gold label. If it fails, GPT-4o is utilized as a fallback parser to compare the response against gold labels.

\textbf{Compliance with harmful instruction: Jailbreak.}\quad
This benchmark tests the model's compliance with harmful instructions, where the urge to satisfy a user's request overrides safety alignment training. We adopt the \emph{forbidden question set} which comprises 390 test cases spanning 13 scenarios with 30 questions each and pair each harmful query with a jailbreak template designed to bypass safety filters.
We generate responses using open-ended sampling with parameters \texttt{temperature=0.7}, \texttt{top\_k=20}, \texttt{top\_p=0.8} and a maximum output length of 256 tokens. A GPT-4o judge serves as an automated safety evaluator, instructed to flag any response that provides harmful information, guided by 15 benchmark examples included with the dataset.

\textbf{Definition of Compliance Rate.}\quad
To enable a comparative analysis across these diverse benchmarks, we define a unified metric, \textit{Compliance Rate}, which quantifies the model's propensity to yield to the prompt's intent. Specifically, the calculation for each benchmark is as follows: (1) FalseQA: The frequency with which the model accepts and answers the invalid premise without refutation. (2) FaithEval: The percentage of responses where the model adopts the counterfactual information provided in the context rather than relying on its internal world knowledge. (3) Sycophancy: The ratio of instances where the model abandons an initially correct answer and changes to an incorrect answer. (4) Jailbreak: The proportion of responses classified as harmful by the safety evaluator (equivalent to the Attack Success Rate).

\subsection{Tracing the Origin of H-Neurons}
Having established the causal role of H-Neurons in instruction-tuned models, a critical question remains unsolved: Are they introduced during post-training alignment phase or already present in the pre-trained phase? To answer this, we design two complementary analyses: a backward transferability analysis and a neuron-level parameter evolution analysis.

% To determine whether H-Neurons arise during the pre-training stage or emerge from post-training alignment, we perform two complementary analyses. First, we assess the backward transferability of H-Neurons by applying detection probes trained on instruction-tuned models directly to their corresponding base models. Second, we examine parameter evolution by comparing neuron-level weight changes before and after alignment. Together, these analyses allow us to identify when H-Neurons first become distinguishable and whether alignment meaningfully modifies them.

\subsubsection{Backward Transferability Analysis}
Our first approach investigates whether the functional distinction between faithful and hallucinatory neurons exists before alignment. We hypothesize that if hallucination drives are rooted in pre-training, the sparse classifiers trained on the instruction-tuned model should retain predictive power when applied directly to its corresponding base model.

\textbf{Standardizing Base Model Decoding.}\quad
Directly comparing base and instruction-tuned models is challenging due to their divergent output formats. Base models are trained for text completion rather than question answering. To ensure a valid comparison, we standardize the decoding process. For each query in TriviaQA, NQ-Open, and BioASQ, we append a strict prompt suffix \textit{"\textbackslash nAnswer:"} and terminate generation upon the first newline character. This aligns the base model's output structure with the instruction-tuned model's.

% We evaluate whether H-Neurons identified in instruction-tuned models retain predictive power in their corresponding base models.

% To ensure consistent answer extraction across model types, we first standardize the decoding setup for base models, which are not instruction-tuned and therefore do not reliably follow question–answer formats. For each question in TriviaQA, NQ-Open, and BioASQ, we append the token string \textit{"\textbackslash nAnswer:"} to the end of every question, and we terminate decoding as soon as the model emits a newline character. All other decoding parameters match those used for instruction-tuned models.

\textbf{Evaluation via Threshold-Invariant Metrics.}\quad
We apply the logistic regression probes derived in Section~\ref{sec:identify_neurons} directly to the base model's activation states without retraining to examine whether the identified H-Neurons exhibit similar activation patterns within the pre-trained models.
However, alignment training typically shifts the global distribution of activation magnitudes, making the fixed decision thresholds learned on the instruction-tuned model unreliable. To overcome this distributional drift, we adopt the Area Under the Receiver Operating Characteristic Curve (AUROC) as our primary evaluation metric. Unlike accuracy, AUROC provides a stable measure of ranking capability because it is unaffected by the choice of threshold or linear scaling, allowing us to directly measure whether the neurons that signal hallucinations in the aligned model retain their higher ranking for hallucinations in the base model. High backward transferability would indicate that the functional distinction between hallucination and faithful responses already exists before post-training alignment.

\subsubsection{Neuron-Level Parameter Evolution}

Our second approach quantifies the physical modifications applied to these neurons during the alignment process. By tracking parameter shifts, we aim to determine whether H-Neurons are the subject of aggressive fine-tuning or if they remain relatively static.

We define the mechanistic drift of a neuron based on the cosine similarity between its weights before and after instruction tuning. Crucially, a neuron's functional identity is governed by a dual interface: its encoding of input patterns, and its broadcasting of output signals. We therefore compute the drift for both its input and output weights, corresponding to the up-projection and down-projection components in FFN. To capture the full scope of functional adaptation, we therefore compute the drift for both its up- and down-projection weights:
\[
\Delta_j^\text{up} = 1 - \cos(W_{\text{up}}^{(j, \text{base})}, W_{\text{up}}^{(j, \text{chat})}),\qquad
\Delta_j^\text{down} = 1 - \cos(W_{\text{down}}^{(j, \text{base})}, W_{\text{down}}^{(j, \text{chat})}).
\]
Larger $\Delta$ values indicate greater modification.
Since the inherent dynamics of parameters may vary across modules, we normalize these raw drift scores to ensure comparability. We calculate the $z$-scores and average the up- and down-projection drifts to obtain a unified final drift $\Delta_j$:
\[
\tilde{\Delta}_j^\text{up} = \frac{\Delta_j^\text{up}-\mu_\text{up}}{\sigma_\text{up}},\qquad
\tilde{\Delta}_j^\text{down} = \frac{\Delta_j^\text{down}-\mu_\text{down}}{\sigma_\text{down}},\qquad
\Delta_j = \frac{\tilde{\Delta}_j^\text{up} + \tilde{\Delta}_j^\text{down}}{2}.
\]
We then analyze the rank distribution of H-Neurons based on $\Delta_j$. A concentration of these neurons in the high-$\Delta_j$ end would suggest that alignment actively constructs or heavily modifies these neurons. Conversely, a uniform distribution or concentration in the low-$\Delta_j$ regime would provide strong evidence that the function of these neurons is largely inherited from pre-training.

\medskip

\bibliographystyle{citation}
\bibliography{custom}

\newpage

%%%%%%%%%%%%%%%%%%%%%%%%%%%%%%%%%%%%%%%%%%%%%%%%%%%%%%%%%%%%

\appendix

\end{document}